\documentclass[twoside,11pt]{article}

\usepackage[accepted]{melba}

\usepackage{amsmath}
\usepackage{xspace}
\usepackage{siunitx}
\usepackage{booktabs}
\usepackage{wrapfig}
\hypersetup{hidelinks}

\DeclareMathOperator*{\argmin}{arg\,min}

\newcommand{\moving}{m}
\newcommand{\moved}{\moving \circ \phi}
\newcommand{\fixed}{f}
\newcommand{\movingseg}{s_m}
\newcommand{\movedseg}{\movingseg \circ \phi}
\newcommand{\fixedseg}{s_f}
\newcommand{\Loss}{\mathcal{L}}
\newcommand{\optimal}{\lambda^*}
\newcommand{\sumpairs}{\sum_{\substack{\moving, \fixed \in \mathcal{D}}}}

\newcommand{\methodname}{HyperMorph\xspace}
\newcommand{\methodpre}{pre-integrative\xspace}
\newcommand{\methodpost}{post-integrative\xspace}
\newcommand{\methodfull}{full-integrative\xspace}
\newcommand{\subpara}[1]{\vspace{0.2cm} \noindent \textbf{#1.}}
\newcommand{\subparaspace}{\vspace{0.2cm} \noindent}

\melbaheading{2022:003}{https://www.melba-journal.org/papers/2022:003.html}{2021}{1-30}{09/2021}{03/2022}{Hoopes, Hoffmann, Greve, Fischl, Guttag, Dalca}{IPMI 2021}{Aasa Feragen, Stefan Sommer, Mads Nielsen, Julia Schnabel}
\ShortHeadings{\methodname}{Hoopes, Hoffmann, Greve, Fischl, Guttag, Dalca}
\firstpageno{1}

\title{Learning the Effect of Registration \\ Hyperparameters with HyperMorph}

\author{\name Andrew Hoopes \email ahoopes@mgh.harvard.edu \\
	\addr Martinos Center for Biomedical Imaging, Massachusetts General Hospital
	\AND
	\name Malte Hoffmann \email mhoffmann@mgh.harvard.edu \\
	\addr Martinos Center for Biomedical Imaging, Massachusetts General Hospital \\
	\addr Department of Radiology, Harvard Medical School
	\AND
	\name Douglas N. Greve \email dgreve@mgh.harvard.edu \\
	\addr Martinos Center for Biomedical Imaging, Massachusetts General Hospital \\
	\addr Department of Radiology, Harvard Medical School
	\AND
	\name Bruce Fischl \email bfischl@mgh.harvard.edu \\
	\addr Martinos Center for Biomedical Imaging, Massachusetts General Hospital \\
	\addr Department of Radiology, Harvard Medical School \\
	\addr Computer Science and Artificial Intelligence Lab, Massachusetts Institute of Technology
	\AND
	\name John Guttag \email guttag@mit.edu \\
	\addr Computer Science and Artificial Intelligence Lab, Massachusetts Institute of Technology
	\AND
	\name Adrian V. Dalca \email adalca@mit.edu \\
	\addr Martinos Center for Biomedical Imaging, Massachusetts General Hospital \\
	\addr Department of Radiology, Harvard Medical School \\
	\addr Computer Science and Artificial Intelligence Lab, Massachusetts Institute of Technology
}

\begin{document}

\maketitle

\begin{abstract}
We introduce \methodname, a framework that facilitates efficient hyperparameter tuning in learning-based deformable image registration. Classical registration algorithms perform an iterative pair-wise optimization to compute a deformation field that aligns two images. Recent learning-based approaches leverage large image datasets to learn a function that rapidly estimates a deformation for a given image pair. In both strategies, the accuracy of the resulting spatial correspondences is strongly influenced by the choice of certain hyperparameter values. However, an effective hyperparameter search consumes substantial time and human effort as it often involves training multiple models for different fixed hyperparameter values and may lead to suboptimal registration. We propose an amortized hyperparameter learning strategy to alleviate this burden by \textit{learning} the impact of hyperparameters on deformation fields. We design a meta network, or hypernetwork, that predicts the parameters of a registration network for input hyperparameters, thereby comprising a single model that generates the optimal deformation field corresponding to given hyperparameter values. This strategy enables fast, high-resolution hyperparameter search at test-time, reducing the inefficiency of traditional approaches while increasing flexibility. We also demonstrate additional benefits of \methodname, including enhanced robustness to model initialization and the ability to rapidly identify optimal hyperparameter values specific to a dataset, image contrast, task, or even anatomical region, all without the need to retrain models. We make our code publicly available at~\url{http://hypermorph.voxelmorph.net}.


\end{abstract}


\section{Introduction}

In deformable image registration, dense correspondences are sought to align two images. Classical iterative registration techniques have been thoroughly studied, leading to mature mathematical frameworks and widely used software packages~\citep{ashburner2007,avants2008,beg2005,fischl1999high,rueckert1999,vercauteren2009}. More recent learning-based registration strategies use image datasets to learn a function that rapidly produces a deformation field for a given image pair~\citep{balakrishnan2019,rohe2017,sokooti2017,de2019,wu2015,yang2017}. These techniques necessitate the tuning of registration hyperparameters that have dramatic impacts on the estimated deformation field. For example, optimal hyperparameter choices can differ substantially across model implementation or even image contrast and anatomy, and even small changes can have large influences on accuracy. Choosing hyperparameter values is therefore an important step in developing, testing, and distributing registration methods.

\begin{figure}[t]
  \centering
  \includegraphics[width=\textwidth]{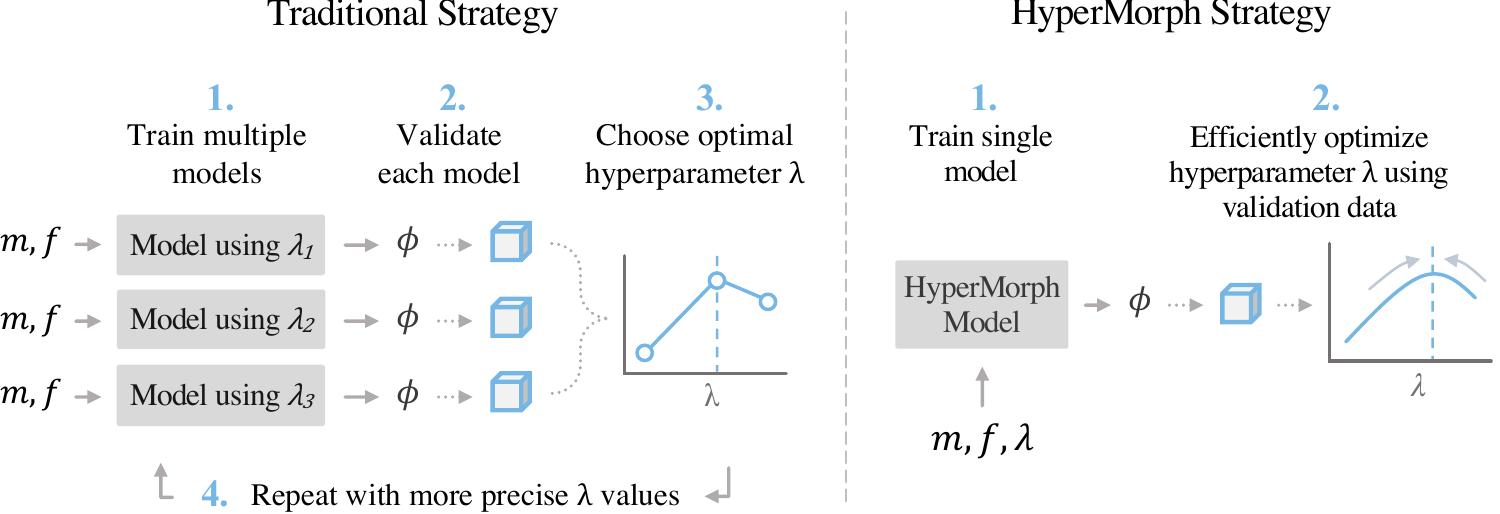}
  \caption{Traditional hyperparameter search strategies (left) involve the optimization of multiple registration models (that predict a deformation~$\phi$ for an input image pair~$m,f$) using different hyperparameter values ($\lambda$) and often require repeating the search for finer hyperparameter resolutions or different ranges. The \methodname strategy (right) trains a \textit{single} network that approximates a landscape of traditional models and can be evaluated for any hyperparameter value at test-time.}
  \label{fig:method-schematic}
\end{figure}

Tuning hyperparameters often involves random or grid search strategies to evaluate separate models for specific discrete hyperparameter values (Figure~\ref{fig:method-schematic}). In practice, researchers or model users typically go through an iterative process of optimizing and validating models using a small subset of hyperparameter values and repeatedly adapting this subset based on the observed results. An optimal value for each hyperparameter is usually selected based on model performance, most often determined by human evaluation or additional validation data, such as anatomical annotations. This approach necessitates considerable computational and human effort, which, in turn, may lead to suboptimal parameter choices, misleading negative results, and impeded progress, especially when researchers resort to using values from the literature that are not appropriate for their specific dataset or registration task. For example, cross-subject registration of neuroimaging data from Alzheimer's Disease patients with significant atrophy will require a substantially different optimal regularization hyperparameter than longitudinal same-subject registration, as we illustrate in our experiments.

We present \methodname, a markedly different strategy for tuning registration hyperparameters. Our contributions are:

\subpara{Method} \methodname involves the end-to-end training of a single, rich model that~\textit{learns} the influence of registration hyperparameters on deformation fields, in contrast to traditional hyperparameter search (Figure~\ref{fig:method-schematic}). A \methodname model comprises a meta network, or a hypernetwork, that estimates a spectrum of registration models by learning a continuous function of the hyperparameters and only needs to be trained once, facilitating rapid image registration for any hyperparameter value at test-time. This avoids the need to repeatedly train a set of models for separate, fixed hyperparameters, since \methodname can correctly predict their outputs in substantially less computational time. Consequently, \methodname facilitates rapid optimization of hyperparameters for a set of validation data. This is even more important for tasks involving more than one important hyperparameter, in which the computational complexity renders traditional search strategies inadequate.

\subpara{Properties} By capitalizing on the similarity of networks with similar hyperparameters, an individual \methodname model employs weight-sharing to optimize efficiently relative to the time required to train the multiple registration models it is able to encompass. Furthermore, we demonstrate that \methodname registration accuracy is less variable across multiple random network initializations compared to conventional registration models, reducing the need to retrain.

\subpara{Utility} \methodname facilitates rapid \textit{test-time} search of optimal hyperparameter values through automatic optimization or visual evaluation for a continuous range of hyperparameters. We show the benefit of this technique by employing a \textit{single} \methodname model to identify optimal hyperparameter values for different loss metrics, datasets, anatomical regions, or tasks with substantially more precision than grid search methods.

\subparaspace
This paper extends work presented at the 2021 International Conference on Information Processing in Medical Imaging~\citep{hoopes2021}. This extension introduces and analyzes an alternative approach to learning the effect of registration hyperparameters by integrating an additional hyperparameter input within monolithic registration networks, as an alternative to using hypernetworks. We contrast the hypernetwork-based \methodname approach with this integrative approach. In addition, we also improve the hypernetwork-based \methodname architecture. In our experiments, we add additional analyses for the effect of network size and hyperparameter sampling strategy on \methodname accuracy, and  evaluate the ability of \methodname to learn the effect of multiple hyperparameters in semi-supervised training using 3D images (as opposed to 2D slices). We also introduce a thorough discussion of this paradigm.


\section{Related Work}

In this section, we introduce the techniques and common hyperparameters involved in modern image registration, and we provide an overview of hyperparameter tuning methods and hypernetwork-based architectures in machine learning.

\subsection{Image Registration}
Image registration is widely studied in many formulations. Classical registration methods find a deformation field by optimizing an energy function independently for each image pair. This often involves maximizing an image-matching term that measures similarity between aligned images while enforcing a regularization on the deformation field to encourage topological correctness or smoothness on the resulting warp. Methods include B-spline based deformations~\citep{rueckert1999}, discrete optimization methods~\citep{dalca2016,glocker2008}, elastic models~\citep{bajcsy1989}, SPM~\citep{ashburner2000}, LDDMM~\citep{beg2005,cao2005,hernandez2009,joshi2000,miller2005,zhang2017}, symmetric normalization~\citep{avants2008}, Demons~\citep{vercauteren2009}, DARTEL~\citep{ashburner2007}, and spherical registration~\citep{fischl1999high}. These techniques are robust and yield precise alignments, but iterative pairwise registration is typically computationally costly, often requiring tens of minutes or more to align image volumes (with size~$256^3$) on a CPU. More recent GPU-based implementations are faster and operate on the order of minutes or even seconds, but require access to a GPU for each registration~\citep{brunn2021,modat2010,shamonin2014}.

Recent learning-based approaches to registration use convolutional neural networks (CNNs) to learn a function that computes the deformation field for a given image pair in seconds on a CPU or faster on a GPU. Supervised models are trained to predict deformation fields that have been been simulated or computed by other techniques~\citep{krebs2017,rohe2017,sokooti2017,yang2017}, whereas unsupervised, or self-supervised, strategies are trained end-to-end and optimize an energy function similar to classical cost functions~\citep{balakrishnan2019,dalca2019varreg,krebs2019,mok2020,de2019,zhao2019}. Semi-supervised strategies leverage auxiliary information, like anatomical annotations, in the loss function to improve test registration accuracy~\citep{balakrishnan2019,hering2019,hoffmann2020,hu2018weakly}.

Commonly, these methods depend on at least one influential hyperparameter that balances the weight of the image-matching term with that of the deformation-regularization term. Semi-supervised losses might require an additional hyperparameter to weight an auxiliary term. Furthermore, the loss terms themselves often contain important hyperparameters, like the number of bins in mutual information~\citep{viola1997alignment} or the neighborhood size (window size) of local normalized cross-correlation~\citep{avants2011}. Unfortunately, tuning hyperparameters in classical registration is an inefficient procedure since it typically requires tens of minutes to hours to compute pair-wise registrations. Although learning-based methods facilitate rapid registration at test-time, training individual models for different hyperparameter values is computationally expensive and can take days or even weeks to converge, resulting in hyperparameter searches that consume hundreds of GPU-hours~\citep{balakrishnan2019,hoffmann2020,de2019}.

\subsection{Hyperparameter Optimization}

Hyperparameter tuning is a fundamental component of general learning-based model development that aims to jointly optimize a validation objective conditioned on model hyperparameters and a training objective conditioned on model weights~\citep{franceschi2018}. In common hyperparameter optimization methods, model training is considered a black-box function. Standard, popular approaches include random, grid, and sequential search~\citep{bergstra2012}. More sample-efficient approaches involve Bayesian optimization techniques, which adopt a probabilistic model of the objective function to seek optimal hyperparameter values~\citep{bergstra2011,mockus1978,snoek2012}. These methods are often time-consuming because they require multiple model optimizations for each assessment of the hyperparameter. Various adaptations of Bayesian strategies improve efficiency by extrapolating model potential from learning curves~\citep{domhan2015,klein2016}, prioritizing resources to promising models with bandit-based methods~\citep{jamieson2016,li2017}, and evaluating cheap approximations of the black-box function of interest~\citep{kandasamy2017}.

Unlike black-box methods, gradient-based hyperparameter tuning strategies compute gradients of the validation error as a function of the hyperparameters by differentiating through the nested learning procedure. Reverse-mode automatic differentiation facilitates the optimization of thousands of hyperparameters, but reversing the entire training procedure is exceedingly memory intensive~\citep{bengio2000,domke2012,maclaurin2015}. To conserve overhead, DrMAD~\citep{fu2016} approximates the training procedure reversal by accounting for the parameter trajectory, and other approaches consider only the last parameter update for each optimization iteration~\citep{luketina2016}. Alternative approaches compute the hyperparameter gradient by deriving an implicit equation for the gradient under certain conditions~\citep{pedregosa2016} or in real-time through forward-mode differentiation~\citep{franceschi2017}.

All of these automatic hyperparameter tuning methods require optimization for an explicit validation objective. However, a comprehensive set of annotated validation data might not be available for every registration task, and in some cases registration accuracy must be evaluated visually or through a non-differentiable downstream measure. Furthermore, hyperparameters are generally optimized once for single set of validation data, and it is not easy to modify hyperparameter values rapidly (e.g.\ for a new task) without retraining models.

\subsection{Hypernetworks}

Hypernetworks are meta neural networks that output the weights of a primary network~\citep{ha2016,schmidhuber1993}, and these two networks comprise a single model that is trained end-to-end. Hypernetworks were originally introduced to compress model size~\citep{ha2016}, but they have been used in a variety of applications across other domains, including posterior estimation in Bayesian neural networks~\citep{krueger2017,ukai2018}, automatic network pruning~\citep{li2020,liu2019}, functional representation~\citep{klocek2019,spurek2020}, multi-task learning~\citep{meyerson2019}, and generative models~\citep{ratzlaff2019}. The influence of hypernetwork initialization strategies has also been explored extensively~\citep{chang2019}.

Additionally, hypernetworks have drawn recent attention as a promising tool for gradient-based hyperparameter optimization, as they facilitate direct differentiation through the entire learning procedure with respect to the hyperparameters of interest. For example, SMASH~\citep{brock2017} employs a hypernetwork to estimate model parameters for a given architecture. Other frameworks use hypernetworks to tune regularization hyperparameters for image classification models and demonstrate that hypernetworks are capable of approximating the overall effect of these hyperparameters~\citep{lorraine2018, mackay2019}. \methodname employs hypernetworks in the context of learning-based registration to learn how hyperparameter values impact predicted deformation fields, similar to recent work for \textit{k}-space reconstruction~\citep{wang2021}. A parallel, independent work also investigates learning the effect of regularization weights in registration models. The proposed method presents a different mechanism that emphasizes conditional instance normalization~\citep{dumoulin2016learned} and employs an MLP, conditioned on the regularization parameter, to shift the feature statistics of each internal feature map~\citep{mok2021}.


\section{Methods}

\subpara{Registration} Deformable registration methods align a moving image~$\moving$ and a fixed image~$\fixed$ by computing a correspondence~$\phi$. We build on unsupervised learning-based registration approaches, which establish a standard registration network~\mbox{$g_{\theta_g}(\moving, \fixed) = \phi$}, with trainable parameters~$\theta_g$, that predicts the optimal deformation~$\phi$ for the input image pair~$\{\moving, \fixed\}$. The deformation map~$\phi$ is often implemented by adding a predicted displacement field to the identity map of the~$n$-dimensional spatial domain~$\Omega \in \mathbb{R}^n$.

These models contain a variety of hyperparameters, and the underlying objective of \methodname is to learn the effect of \textit{loss} hyperparameters~$\Lambda$ on the deformation field~$\phi$. We propose two fundamentally different ways of achieving this. In the first, we employ a hypernetwork to modify the registration function~$g_{\theta_g}$ as a function of hyperparameters~$\Lambda$. In the second, we extend the existing registration function~$g_{\theta_g}$ to take in hyperparameters~$\Lambda$ as input. We focus our development on the former, hypernetwork-based approach, which is substantially easier to optimize and yields better results.

\subsection{\methodname}

\begin{figure}[t]
\centering
  \includegraphics[width=\textwidth]{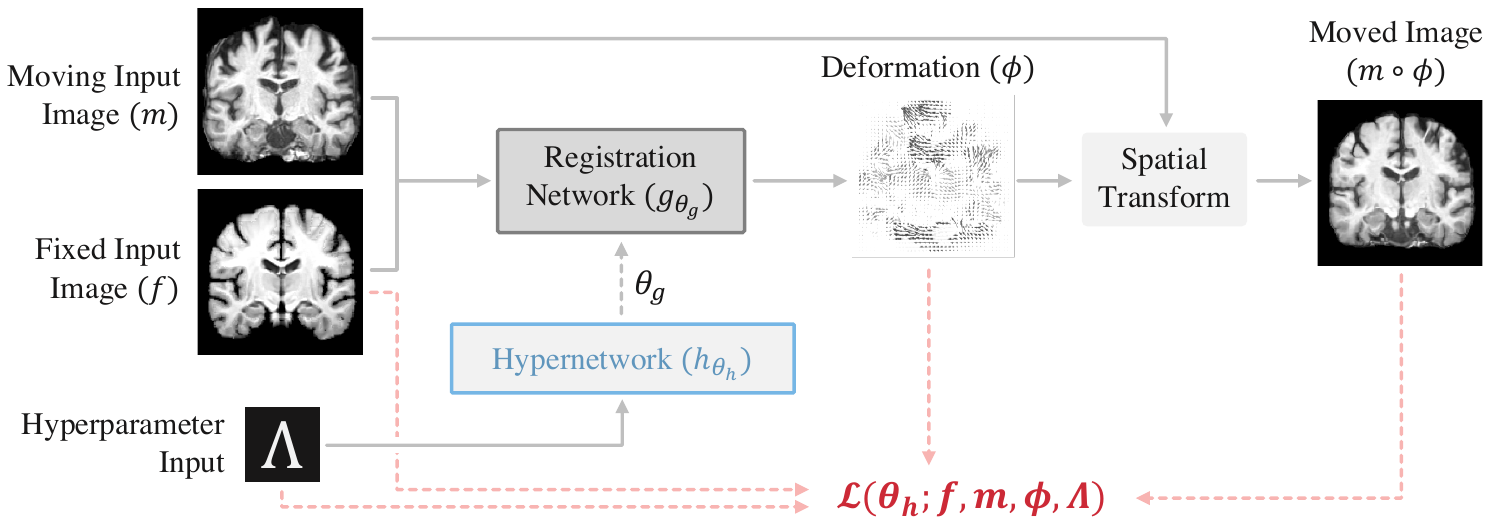}
  \caption{The \methodname architecture comprises a hypernetwork~$h_{\theta_h}(\Lambda) = \theta_g$ that takes registration hyperparameters~$\Lambda$ as input and estimates the parameters of a primary registration network~$g_{\theta_g}$. \methodname is trained end-to-end as a single model with a continuous range of hyperparameter values, capitalizing on the implicit weight-sharing that captures the redundancy that exists amongst a landscape of registration networks.
  }
  \label{fig:architecture}
\end{figure}

We propose a nested registration function, in which a hypernetwork~$h_{\theta_h}(\Lambda) = \theta_g$, with parameters~$\theta_h$, estimates the parameters of the primary registration network~$\theta_g$ for input sample values of~$\Lambda$ (Figure \ref{fig:architecture}). We use stochastic gradient methods to optimize hypernetwork parameters~$\theta_h$ with the loss function:
\begin{alignat}{3}
\Loss_h(\theta_h; \mathcal{D})
&=& \mathbb{E}_{\Lambda \sim p(\Lambda)} \Big[ \Loss(\theta_h ; \mathcal{D}, &\Lambda)& \Big] ,
\label{eq:hyper-loss}
\end{alignat}
where~$h_{\theta_h}(\Lambda) = \theta_g$,~$\mathcal{D}$ is a training dataset of images, ~$\Loss(\cdot)$ is a registration loss function with hyperparameters~$\Lambda$, and ~$p(\Lambda)$ is a prior probability over the hyperparameters of interest. This distribution~$p(\Lambda)$ can be uniform over a defined range or tailored to match assumptions. For each optimization step, we sample values from~$p(\Lambda)$ and use these in the loss function~$\Loss(\cdot)$ and as input to the network~$h_{\theta_h}(\cdot)$. Introducing a level of abstraction, the hypernetwork~$h_{\theta_h}$ allows the convolutional kernels~$\theta_g$ of the registration network~$g_{\theta_g}$ to flexibly adapt to varying hyperparameter values.

\subpara{Unsupervised Model Instantiation}
We build on unsupervised approaches to learning-based registration, which commonly involve optimizing a loss of the form:
\begin{align}
\mathcal{L}(\moving, \fixed, \phi; \Lambda) =
\Loss_{sim}(\fixed, \moved; \lambda_{sim}) + \lambda \Loss_{reg}(\phi; \lambda_{reg})  \label{eq:hyper-loss-no-norm}
\end{align}
where~$\moved$ represents~$\moving$ warped by~$\phi$. The~$\Loss_{sim}$ loss term quantifies the similarity between~$\moved$ and~$\fixed$ and includes potential hyperparameters~$\lambda_{sim}$, whereas the~$\Loss_{reg}$ term measures the spatial regularity of the estimated deformation field~$\phi$ and includes potential hyperparameters~$\lambda_{reg}$. The hyperparameter~$\lambda$ regulates the weight of~$\Loss_{reg}$, and we define~$\Lambda = \{\lambda, \lambda_{sim}, \lambda_{reg}\}$. One limitation of this formulation is that~$p(\lambda)$ is challenging to define as the range of $\lambda$ is infinite. We constrain~$\lambda$ to~$[0, 1]$ by scaling~$\Loss_{sim}$ by~$(1 - \lambda)$. We thus optimize \methodname using:
\begin{align}
\Loss_h(\theta_h; \mathcal{D})
= \mathbb{E}_\Lambda \Big[ 
\sumpairs \Big(
(1 - \lambda) \Loss_{sim}(\fixed, \moved; \lambda_{sim}) + \lambda \Loss_{reg}(\phi; \lambda_{reg}) \Big) \Big], \hfill \quad
\raisetag{2\normalbaselineskip}
\label{eq:hyper-loss-classical}
\end{align}
where~$\phi=g_{\theta_g}(\moving, \fixed)$ and~$\theta_g = h_{\theta_h}(\Lambda)$. 

In our experiments, we use mean-squared error (MSE) and \textit{local} normalized cross-correlation (NCC) as the similarity metrics for~$\Loss_{sim}$ when registering images of the same contrast, and we use mutual information (MI) for multi-contrast registration. Local NCC involves a hyperparameter that defines the local neighborhood (window) size, and MI involves a hyperparameter that controls the number of histogram bins~\citep{viola1997alignment}. In some cases, MSE is scaled by estimated image noise~$\sigma^{-2}$.

To encourage diffeomorphic deformations, which are invertible by design, we spatially integrate the vectors of a stationary velocity field (SVF) $v$ using~\textit{scaling and squaring}~\citep{arsigny2006log,ashburner2007,dalca2019varreg} to obtain~$\phi$, which is regularized using
\begin{align}
\Loss_{reg}(\phi) = \frac{1}{2} \sum_{\substack{i=1}}^{n}\|\nabla v_{i}\|^2,
\label{eq:loss-reg}
\end{align}
where~$i$ is an axis in the~$n$-dimensional image and~$\nabla v_i(p)$ defines the spatial gradient of~$v_i$ at location~$p \in \Omega$. The regularization term~$\Loss_{reg}$ can take a variety of forms and might include multiple specific hyperparameters~$\lambda_{reg}$.

\subpara{Semi-supervised Model Instantiation}
Following recent strategies that exploit supplemental information during training, we extend \methodname to semi-supervised learning by incorporating segmentation maps in the loss function:
\begin{alignat}{3}
\Loss_h(\theta_h; \mathcal{D}) = \mathbb{E}_\Lambda 
\sumpairs  &\Big[ 
(1-\lambda)(1-\gamma) \Loss_{sim}(\fixed, \moved; \lambda_{sim})  \Big. \nonumber\\[-10pt] 
& \Big. + \lambda \Loss_{reg}(\phi; \lambda_{reg}) + \gamma \Loss_{seg}(\fixedseg, \movedseg)  \Big] ,
\label{eq:hyper-loss-semisupervised}
\end{alignat}
where~$\movingseg$ and~$\fixedseg$ are segmentation maps corresponding to the moving and fixed images, respectively, and~$\Loss_{seg}$ is a measure of segmentation overlap, often the Dice coefficient~\citep{dice1945}, weighted by the hyperparameter~$\gamma$. As with the unsupervised loss, we constrain the range of~$\gamma$ within~$[0,1]$ by scaling the similarity term~$\Loss_{sim}$ by~$(1-\lambda)(1-\gamma)$.

\subsection{Hyperparameter Tuning}

An optimized \methodname model can rapidly register a test image pair~$\{\moving, \fixed\}$ as a function of important hyperparameters. If external annotation data is not available, hyperparameters may be efficiently tuned in an interactive fashion. In some cases, landmarks, functional data, or segmentation maps are present, facilitating fast automatic hyperparameter optimization for a validation dataset.

\subpara{Interactive} Users can manually adjust hyperparameter values in close to real-time using interactive sliders until they are visually satisfied with the alignment of a given image pair. Sometimes, the user might adopt different settings when focusing on particular domains of the image. For instance, the optimal value of the~$\lambda$ hyperparameter, which balances image-similarity and regularization, can differ substantially across anatomical regions of the brain (see Figure~\ref{fig:downstream-optimization}). Interactive tuning is feasible since \methodname can efficiently estimate the influence of~$\lambda$ values on the deformation field~$\phi$ without necessitating further training.

\subpara{Automatic} If additional information, such as segmentation maps~$\{\movingseg,\fixedseg\}$, are present for validation, an individual trained \methodname model facilitates rapid optimization of hyperparameter values using
\begin{align}
\Lambda^* = \argmin_\Lambda \Loss(\Lambda; \theta_h, \mathcal{D}, \mathcal{V}) 
= \argmin_\Lambda
\sum_{\substack{\moving,\fixed \in \mathcal{D} \\ \movingseg, \fixedseg \in \mathcal{V}}}
\Loss_{val}(\fixedseg, \movingseg \circ \phi), 
\label{eq:val-hyper-loss}
\end{align}
where~$\mathcal{V}$ is a set of validation segmentations and~$\Loss_{val}$ is a validation loss to be minimized. To carry out this hyperparameter optimization, we freeze the hypernetwork parameters~$\theta_h$ so that the input~$\Lambda$ represents the sole set of trainable parameters. We rapidly optimize~\eqref{eq:val-hyper-loss} using stochastic gradient descent strategies.

\subsection{Implementation}
\label{sec:implementation}

We implement \methodname with the open-source VoxelMorph registration library~\citep{balakrishnan2019}, modeling the base registration network~$g_{\theta_g}$ with a U-Net-like architecture~\citep{ronneberger2015}. In our experiments, this comprises a four-layer convolutional encoder-like part, with 16, 32, 32, and 32 respective channels per layer, followed by a seven-layer convolutional decoder-like part, with 32, 32, 32, 32, 32, 16, and 16 respective channels per layer. The convolutional layers have a kernel size of 3, a stride of 1, and are activated using LeakyReLU with~$\alpha$ parameter 0.2. After each convolution in the encoder, we reduce the spatial dimensions using max pooling with a window size of 2, and in the decoder, each convolution is followed by an upsampling layer until the volume is returned to full resolution. Skip connections concatenate features of the encoder with features of the first decoder layer of equal resolution. A final, linearly-activated convolutional layer outputs an SVF, which is integrated with five scaling and squaring steps to obtain~$\phi$~\citep{ashburner2007,dalca2019varreg}. In total, this base model~$g_{\theta_g}$ contains 313,507 trainable parameters.

In the hypernetwork-based \methodname models used throughout our experiments,~$h_{\theta_h}$ consists of five fully-connected layers, each with 32, 64, 64, 128, and 128 respective units and ReLu activations, followed by a final linearly-activated layer with output units corresponding to the number of trainable parameters in~$g_{\theta_g}$. This is improved from the previous \methodname implementation~\citep{hoopes2021}, which yielded slightly worse accuracy compared to some baseline models and used a hypernetwork consisting of four fully-connected layers, each with 64 units and a tanh-activated final layer. Together, the registration network and hypernetwork constitute a single network with approximately~$40.5$ million trainable parameters~$(\theta_h)$ that exist entirely in the hypernetwork. Since the large majority of trainable parameters exist in the final layer of the hypernetwork, the model size increases substantially with the number of parameters in~$g_{\theta_g}$, but this increase does not lead to substantial memory footprint, as this is dominated by the convolutional tensors. We emphasize that the proposed strategy pertains to any learning-based registration architecture, not just VoxelMorph.

We train all \methodname and baseline models with the Adam optimizer~\citep{kingma2014adam}, a batch size of 1, and an initial learning rate of~$10^{-4}$, employing a decay strategy that reduces the learning rate by a factor of two for every~\num{5e4} optimization steps without improvement in the training loss. Continuous hyperparameter values are randomly sampled from a uniform distribution during training. Based on our experiments, the agreement of \methodname with baselines at the boundary hyperparameter values~$\{0,1\}$ of~$\lambda$ can be improved if values are slightly over-sampled during training. We let~$r$ be the fraction of hyperparameters sampled from this end-point distribution~$\lambda\in\{0,1\}$ and set it to 0.2 in our experiments. Discrete hyperparameters, like the local NCC window size, are sampled from a~\textit{discrete} uniform distribution during training, and we normalize the sampled values in this range to~$[0,1]$ when used as input to the hypernetwork. We observe that \methodname learns a continuous function for these hyperparameters by interpolating weights across discrete values, enabling their direct optimization at test-time using gradient strategies. We implement \methodname in Python, using the TensorFlow~\citep{abadi2016} and Keras~\citep{chollet2015} packages, and release \methodname as a component of the broader VoxelMorph registration package, with plans to support a PyTorch~\citep{paszke2019pytorch} implementation. We train and evaluate all models on Nvidia Quadro RTX 8000 GPUs.

\begin{figure}[t]
\centering
  \includegraphics[width=\textwidth]{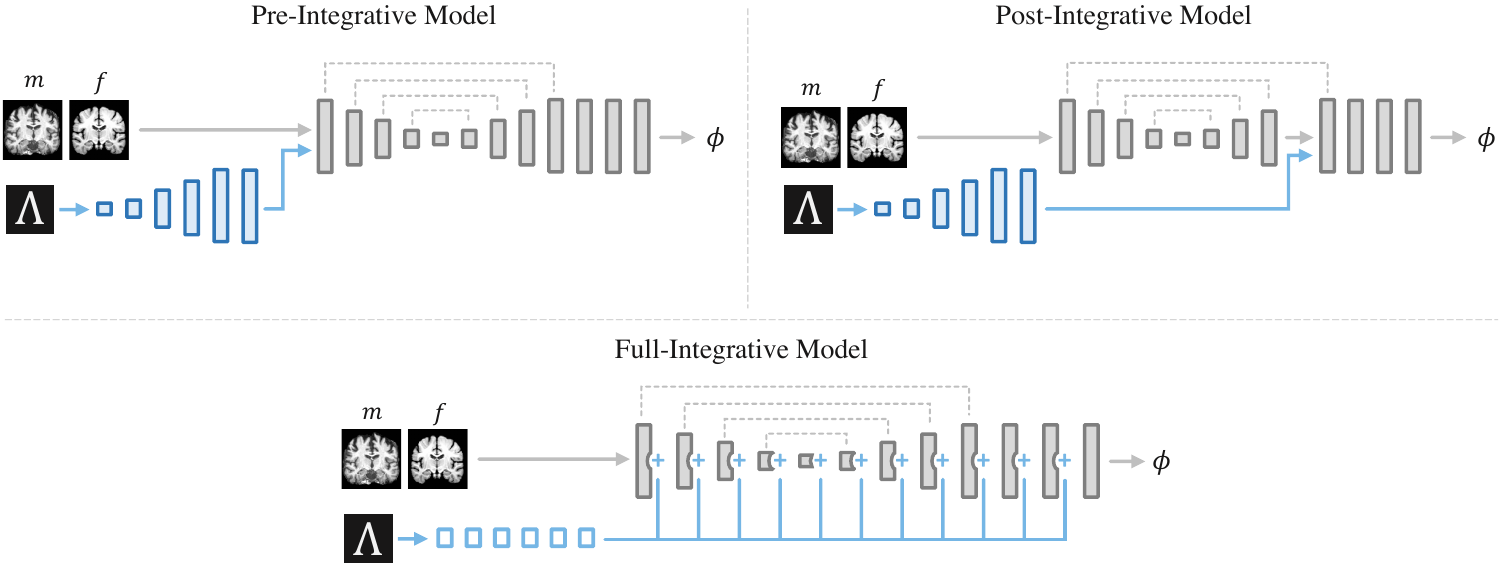}
  \caption{Alternative model architectures for learning the effect of registration hyperparameters. In these approaches, the hyperparameters~$\Lambda$ are provided as input to an auxiliary convolutional network~$d_{\theta_d}$ (blue), which is integrated directly with the primary registration network~$g_{\theta_g}$ (grey). The output of~$d_{\theta_d}$ is either added to the output channels of the registration U-Net (full-integrative model) or provided as an additional input to the first layer (pre-integrative model) or last upsampling layer (post-integrative model).}
  \label{fig:alt-architecture}
\end{figure}

\subsection{Alternative Models}
\label{sec:altmethod}
We also analyze a fundamentally different approach to amortized hyperparameter learning by extending the \textit{inputs} to the registration function as opposed to changing the registration function using hypernetworks. We build on architectures that combine scalar or non-image inputs with convolutional networks used in other tasks, such as probabilistic segmentation~\citep{kohl2018} or conditional template construction~\citep{dalca2019templates}.

We examine three alternative implementations, in which hyperparameters are provided as input to a small auxiliary convolutional sub-network~$d_{\theta_d}$, with parameters~${\theta_d}$, that is joined directly with the primary registration network (Figure~\ref{fig:alt-architecture}). In the first two alternative architectures, input hyperparameters are repeated and reshaped to an~$8\times8\times8$ multi-channel volume, with one channel for each input hyperparameter, and provided as input to a series of six convolutional layers in~$d_{\theta_d}$, each with 32 channels. The output of each layer in~$d_{\theta_d}$ is upsampled until the target image resolution is reached. In the first alternative architecture, the \methodpre network,~$g_{\theta_g}$ takes the output of~$d_{\theta_d}$ as an additional input~\citep{dalca2019templates}. In the second architecture, the \methodpost network, the output of~$d_{\theta_d}$ is concatenated with the input to the \textit{final} upsampling layer of the U-Net~\citep{kohl2018}. In the third alternative architecture,~$d_{\theta_d}$ comprises five fully-connected layers, each with 256 units and ReLu activations, followed by a linearly-activated layer with output units equal to the total number of channels across all layers in~$g_{\theta_g}$. We refer to this architecture as the \methodfull network, and each value estimated by~$d_{\theta_d}$ is added to its corresponding convolutional output channel in the base network.


\section{Experiments}

We conduct experiments evaluating how well a single \methodname model captures the behavior and matches the performance of individually trained registration networks with separate hyperparameter values. We show that our approach substantially reduces the computational and human effort required for a search with one or two registration hyperparameters. We present considerable improvements in robustness to model initialization. We also illustrate the utility of \methodname for efficient hyperparameter optimization across different subpopulations of data, image contrasts, registration types, and individual anatomical structures. Additionally, we compare hypernetwork-based \methodname with the proposed alternative models that expand the input space, and we provide a framework analysis exploring the effect of hypernetwork size and hyperparameter sampling on \methodname performance.

\begin{table}
\centering
\caption{Three groups of image datasets are used throughout the experiments and split into train, validate, and test subsets of specified size.}
\begin{tabular}{l|c|c|c|l}
\toprule
\multicolumn{1}{c|}{\textbf{Group}} & \textbf{Train} & \textbf{Validate} & \textbf{Test} & \multicolumn{1}{c}{\textbf{Datasets}} \\ \hline
Within-contrast       & 7,400 & 5,000 & 5,030 & ABIDE~\citep{di2014}  \\
                      &     &     &     & ADHD-200~\citep{milham2012}  \\
                      &     &     &     & GSP~\citep{dagley2017}  \\
                      &     &     &     & MCIC~\citep{gollub2013}  \\
                      &     &     &     & OASIS-1~\citep{marcus2007}  \\
                      &     &     &     & PPMI~\citep{marek2011}  \\
                      &     &     &     & UK Biobank~\citep{sudlow2015}  \\
                      &     &     &     & Buckner40~\citep{fischl2002}  \\
\hline
Multi-contrast        & 496 & 496 & 496 & HCP~\citep{bookheimer2019} \\
                      &     &     &     & FSM (in-house data) \\
\hline
Longitudinal          & 48 & 48 & 48 & OASIS-2~\citep{marcus2010}  \\
\bottomrule
\end{tabular}
\label{tab:datasets-table}
\end{table}

\subpara{Datasets}
We use three groups of 3D brain magnetic resonance imaging~(MRI) data gathered across multiple public datasets, as summarized in Table~\ref{tab:datasets-table}. The first group includes a series of within-contrast T1-weighted~(T1w) scans, and the second group is a multi-contrast collection of T1w and T2-weighted~(T2w) images, FLASH scans acquired with various flip angles, and MPRAGE scans with different inversion times. We also employ a group of longitudinal images for comparisons between within-subject and cross-subject registration tasks, in which we consider two T1w scans, acquired at least one year apart for each individual.

Using FreeSurfer 7.2~\citep{fischl2012}, all MR images are resampled as 256$\times$256$\times$256 volumes with 1-mm isotropic resolution, bias-corrected, brain-extracted, and automatically anatomically segmented for evaluation. We affinely align all images to the FreeSurfer Talairach atlas and uniformly crop them to size 160$\times$192$\times$224. When evaluating registration accuracy with segmentation data, we consider standard anatomical labels provided by FreeSurfer: the thalamus, caudate, putamen, pallidum, hippocampus, amygdala, accumbens area, ventral diencephalon, choroid plexus, cerebral cortex, cerebral white matter, cerebellar cortex, cerebellar white matter, brain stem, cerebrospinal fluid, and the 3rd, 4th, and lateral ventricles. FreeSurfer generates accurate segmentations that are often considered a silver-standard for automatic brain labeling~\citep{dalca2019unsupervised,puonti2016}, but we also employ an auxiliary set of 30 manually-labeled T1w images from the Buckner40 cohort to evaluate registration accuracy using gold-standard annotations. This dataset was not used during training.

\begin{figure}[t]
  \centering
  \includegraphics[width=\textwidth]{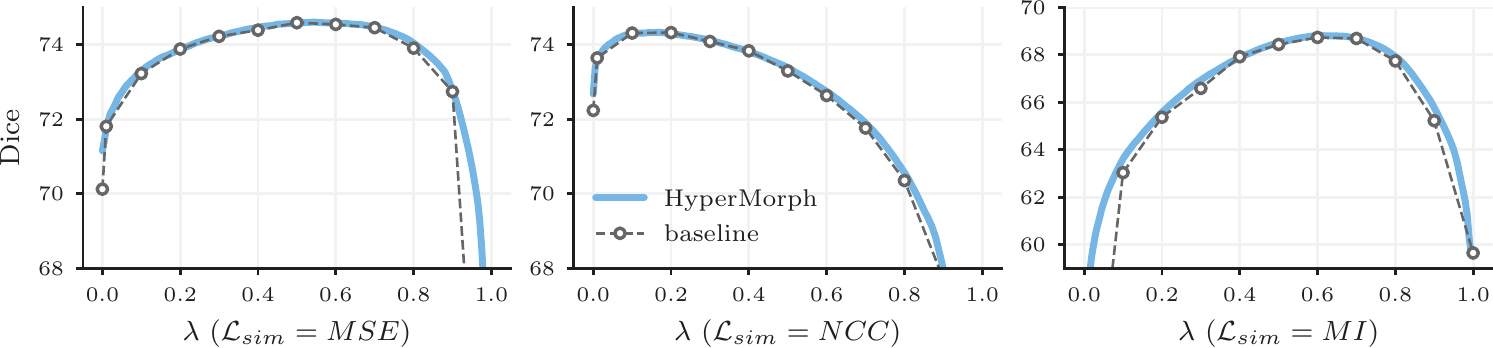}
  \caption{Mean Dice scores achieved by a single \methodname model (blue) and baselines trained for different regularization weights~$\lambda$ (grey) when using MSE, NCC, or MI similarity metrics.
  }
  \label{fig:lambda-baseline-comparison}
\end{figure}

\subpara{Evaluation metrics}
For evaluation, we compute the volumetric Dice overlap coefficient (reported as percentage points between 0 and 100) as well as the 95th percentile surface distance in millimeters for corresponding anatomical labels of the moved and fixed segmentation maps. To quantify regularity of the deformation~$\phi$, we report the standard deviation of the Jacobian determinant~$|J_{\phi}|$, where~$J_{\phi}(p) = \nabla \phi(p)$ for each displacement voxel~$p \in \Omega$.

\subpara{Baseline Models}
\methodname can be applied to any learning-based registration architecture. To analyze how accurately it captures the effect of hyperparameters on the inner registration network~$g_{\theta_g}(\cdot)$, we train baseline VoxelMorph models with architectures identical to~$g_{\theta_g}(\cdot)$, each with a different set of fixed hyperparameter values.

\subsection{Experiment 1: \methodname Efficiency and Capacity}

The goal of this experiment is to assess the extent to which a single \methodname model captures a landscape of baseline models trained with different hyperparameter values. We emphasize that we do not focus on comparing \methodname with the latest registration architecture but rather on evaluating how \methodname can be combined with any framework.

\begin{figure}[t]
  \centering
  \includegraphics[width=\textwidth]{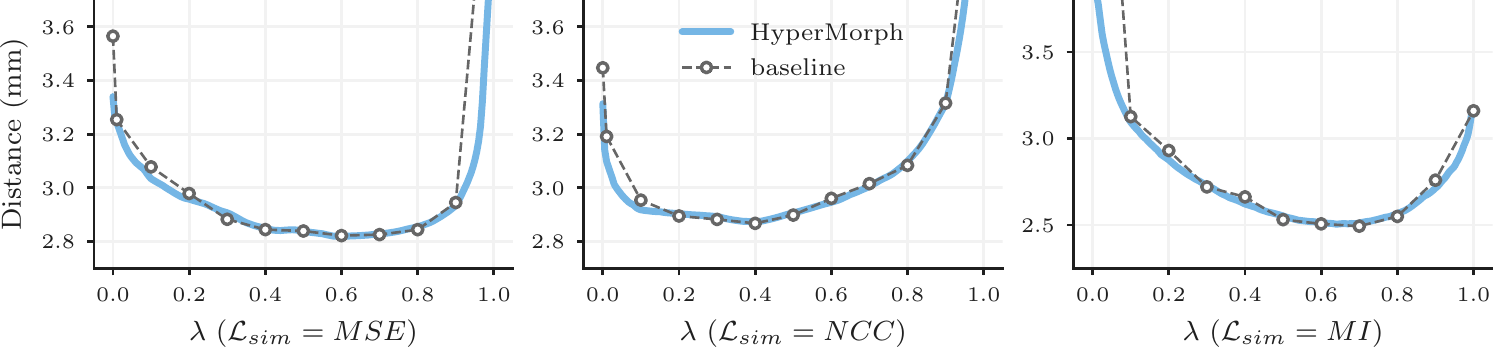}
  \caption{Mean 95th percentile surface distances achieved by a single \methodname model (blue) and baselines trained for different regularization weights~$\lambda$ (grey) when using MSE, NCC, or MI similarity metrics.}
  \label{fig:dist-baseline-comparison}
\end{figure}

\subpara{Setup}
We first compare the accuracy and computational cost of a single \methodname model to standard grid hyperparameter search for the regularization weight~$\lambda$. In separate analyses, we train \methodname and the VoxelMorph baselines using the MSE ($\sigma = 0.05$) and NCC (window size $= 9^3$) similarity metrics for within-contrast registration, as well as the MI metric (32 fixed bins) for cross-contrast registration. For each metric, we train 12 baseline models and compare network performance across 50 randomly selected image pairs from the test set. To analyze \methodname in the context of domain-shift scenarios, we further evaluate models (trained with~$\Loss_{sim} = MSE$) on 20 image pairs from the manually-labeled Buckner40 cohort, held out entirely from training.

Additionally, we assess the ability of \methodname to learn the effect of multiple hyperparameters simultaneously. First, we train a semi-supervised \methodname model using a subset of six labels, holding out a further six labels for evaluation, to simulate partially annotated data. In this experiment, we choose~$\lambda$ and the relative weight~$\gamma$ of the semi-supervised loss~\eqref{eq:hyper-loss-semisupervised} as the hyperparameters of interest. Second, we train a \methodname model treating~$\lambda$ and the local NCC window size~$w$ as hyperparameters. Since the computation of local NCC is computationally prohibitive for large window sizes in 3D data, we conduct the experiment in 2D on mid-coronal slices. These slices are not bias-corrected during preprocessing, since the local NCC metric is most useful for aligning images with strong intensity inhomogeneities. We train semi-supervised baseline models for 25 hyperparameter combinations, performing a discrete search on a~$5\times5$ two-dimensional grid.

\begin{table}[b]
\centering
\caption{Total train time (left) and model variability across random initializations (right) for \methodname and baseline grid search techniques. Train time for the 2 hyp.~($\lambda,w$) experiment is substantially faster as it was conducted using 2D image slices as opposed to 3D volumes.}
\begin{tabular}{l|ccc|cc}
\toprule
 & \multicolumn{3}{c|}{\textbf{Train time (total GPU-hours)}} & \multicolumn{2}{c}{\textbf{Variability (SD)}} \\
 & 1 hyp. ($\lambda$) & 2 hyp. ($\lambda,\gamma$) & 2 hyp. ($\lambda,w$) & MSE & MI \\ \hline
\methodname & \textbf{192.5~$\pm$ 23.1}  & \textbf{321.9~$\pm$ 16.1}  & \textbf{4.0~$\pm$ 0.1}    & \textbf{0.100} & \textbf{0.127} \\
Baseline    & 1,174.9~$\pm$ 196.1        & 4,120.5~$\pm$ 295.4        & 46.8~$\pm$ 5.7            & 0.176          & 0.325         \\
\bottomrule
\end{tabular}
\label{tab:dice-runtime-table}
\end{table}

\begin{figure}[t]
  \centering
  \includegraphics[width=\textwidth]{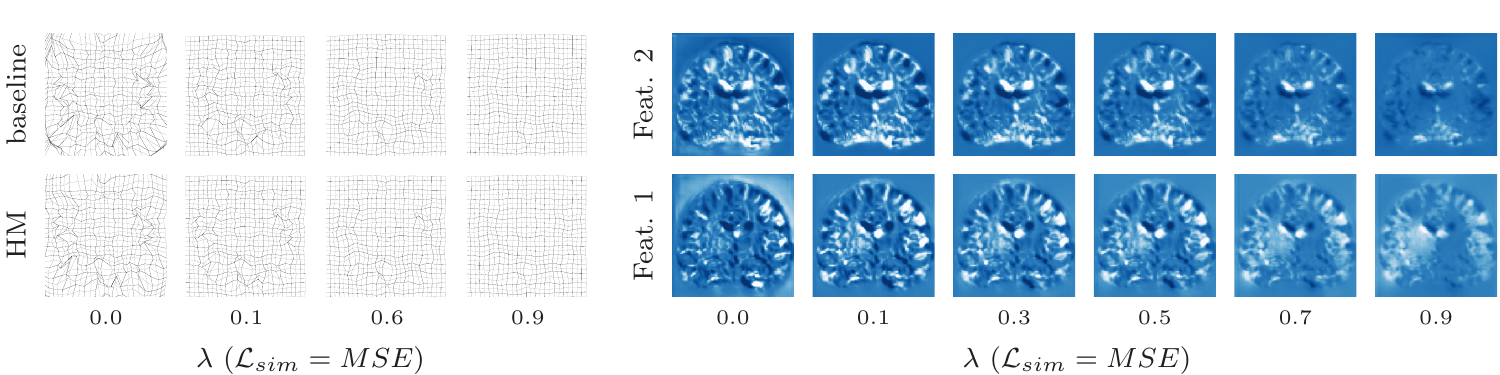}
  \caption{Left: visual comparison of \methodname (HM) and baseline model registration deformations on a mesh grid, illustrating similar changes in regularity across~$\lambda$ values. Right: representative changes in feature activations of the final layer in the HyperMorph U-Net for different regularization weights.}
  \label{fig:supplementary-baseline-comparison}
\end{figure}

\subpara{Results}
\textit{Computational Cost}. A single \methodname model converges considerably faster than the baseline grid search. For single-hyperparameter tests, \methodname requires~$6.1$ times fewer GPU-hours than the 1D grid search with 12 baseline models (Table~\ref{tab:dice-runtime-table}). For two hyperparameters, the difference is even more striking, with \methodname requiring~$12.3$ times fewer GPU-hours than a grid search with 25 baseline models. Furthermore, a 5$\times$5 grid search is coarse, especially if the scale of the evaluated hyperparameters is unknown. While the time required for grid search is proportional to the number of grid points, \methodname enables arbitrarily fine resolution between grid points, at no increase in training time.

\begin{figure}[t]
  \centering
  \includegraphics[width=\textwidth]{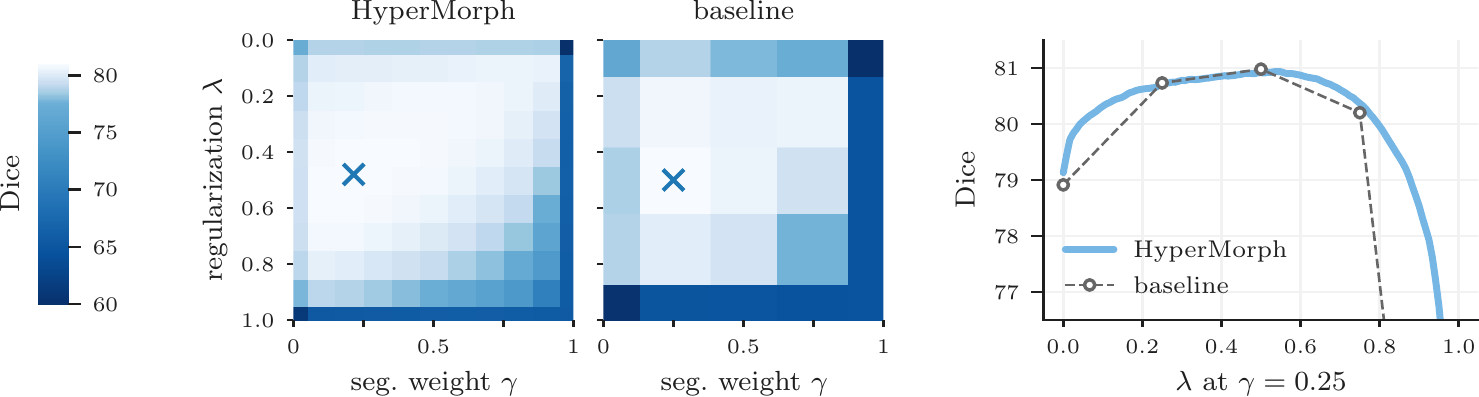}
  \caption{Two-dimensional hyperparameter search for semi-supervised registration with regularization hyperparameter~$\lambda$ and segmentation weight~$\gamma$. For a set of 50 test pairs, the cross markers indicate the optimal~$\lambda,\gamma$ values determined by \methodname and a baseline grid search. We compute total Dice using both sets of training and held-out labels. While the left hyperparameter space is illustrated on a discrete grid for visualization, \methodname enables evaluating the effect of hyperparameter values at arbitrarily fine resolution.}
  \label{fig:semisupervised-baseline-comparison}
\end{figure}

\begin{figure}[t]
  \centering
  \includegraphics[width=\textwidth]{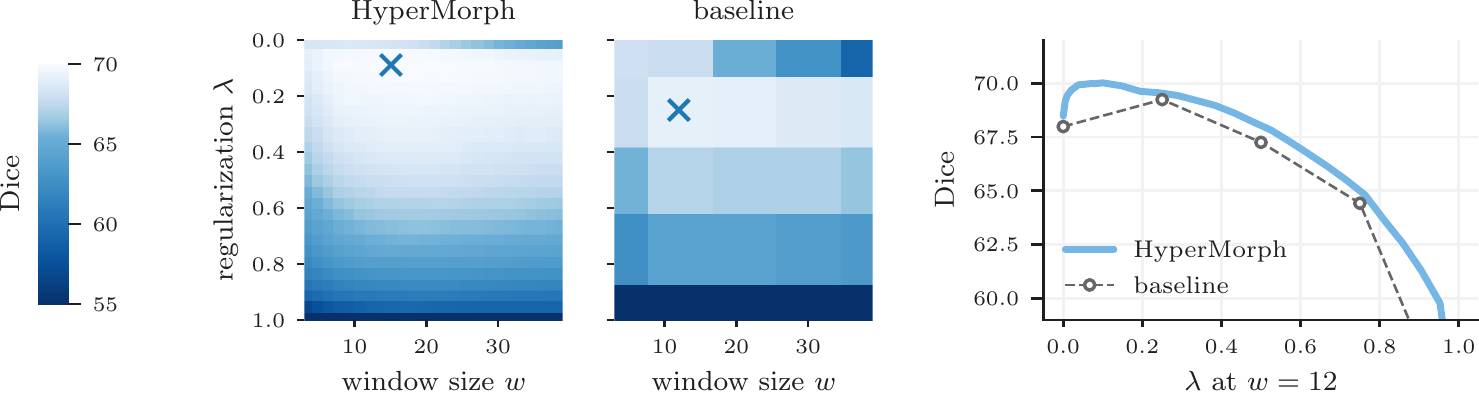}
  \caption{Two-dimensional hyperparameter search for unsupervised registration with regularization weight~$\lambda$ and local NCC window size~$w$. For a set of 50 test pairs, the cross markers indicate the optimal~$\lambda,w$ values determined by \methodname and a baseline grid search. \methodname is able to identify the optimal~$\lambda,w$ value pair missed by a traditional grid search.}
  \label{fig:NCC-window-size-baseline-comparison}
\end{figure}

\textit{Representation accuracy}. Along with the computational advantage, Figures \ref{fig:lambda-baseline-comparison}, \ref{fig:dist-baseline-comparison}, \ref{fig:semisupervised-baseline-comparison}, and \ref{fig:NCC-window-size-baseline-comparison} show that \methodname yields optimal hyperparameter values similar to those determined through the baseline-model grid search. For each image pair, an average difference in the optimal hyperparameter value~$\optimal$ of only~$0.04 \pm 0.06$ across single-hyperparameter experiments results in a negligible maximum Dice difference of~$0.06 \pm 0.42$ (on a scale of~$0$ to~$100$) and a minimum surface distance of~$0.01 \pm 0.02$ millimeters. Even when evaluated on the held-out, manually-labeled dataset, \methodname similarly matches the baseline registration accuracy, differing in maximum Dice by~$0.02 \pm 0.09$ and in minimum surface distance by~$0.01 \pm 0.03$ millimeters (Figure \ref{fig:held-out}). Furthermore, the deformation field regularity at~$\optimal$, measured by standard deviation of the Jacobian determinant, is~$0.31 \pm 0.14$ and differs by only~$0.01 \pm 0.01$ across HyperMorph and baseline models. We visualize these deformation fields and \methodname channel activations in Figures \ref{fig:supplementary-baseline-comparison} and \ref{fig:supplementary-features}.

Semi-supervised experiments yield a maximum Dice difference of only~$0.02 \pm 0.27$ and minimum surface distance of~$0.01 \pm0.01$. Figure \ref{fig:NCC-window-size-baseline-comparison} showcases an example in which the optimal pair of~$\{\lambda,w\}$ values identified by \methodname lies far from the points of the coarse search grid, resulting in a~$0.78 \pm 0.98$ decrease in maximum Dice for the traditional approach. In practice, even fewer baselines might be trained for a coarser hyperparameter search, resulting in either suboptimal hyperparameter choice or sequential search with substantial manual overhead.

\subsection{Experiment 2: Robustness to Initialization}

The goal of this experiment is to analyze the robustness of each hyperparameter search strategy to different network weight initialization.

\subpara{Setup} We repeat the previous single-hyperparameter experiment with MSE and MI, retraining five \methodname models from scratch. For each of four different~$\lambda$ values, we also train five baseline models. Each training run re-initializes the kernel weights with a different randomization seed, and we compare the variability across initializations in terms of the standard deviation~(SD) of Dice accuracy for the \methodname and baseline networks, in a set of 50 image pairs. 

\begin{figure}[t]
  \centering
  \includegraphics[width=0.85\textwidth]{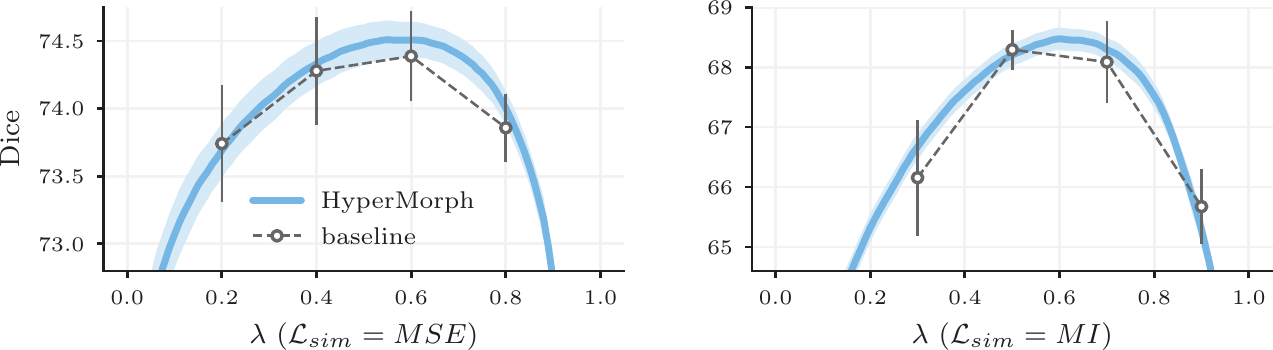}
  \caption{Variability across several training initializations for \methodname and baseline models. Error bars and fill regions span a~2-$\sigma$ range around the mean registration accuracy, which is substantially tighter for \methodname.}
  \label{fig:robustness}
\end{figure}

\subpara{Results} Figure~\ref{fig:robustness} shows that \methodname is considerably more robust (lower SD) to differential initialization than the baselines. Across the entire range of~$\lambda$, the average Dice SD for \methodname models trained with MSE is~$1.7$ times lower ($P<.001$ via paired \textit{t}-test) than the baseline SD and~$2.6$ times lower for MI ($P<.001$) (Table \ref{tab:dice-runtime-table}).

\subsection{Experiment 3: Hyperparameter-Tuning Utility}
\label{sec:utility}

This experiment aims to validate \methodname as a powerful tool for hyperparameter tuning across a number of registration tasks, with or without annotated validation data.

\subpara{Setup}
\textit{Interactive Tuning}. We demonstrate the utility of \methodname through an interactive tool that enables visual optimization of hyperparameters even if no annotated data are available. The user can explore the effect of \textit{continuously varying} hyperparameter values using a single trained model and manually select a preferred optimal deformation. We provide an interactive \methodname demonstration with associated code at~\url{http://hypermorph.voxelmorph.net}.

\textit{Automatic Tuning}. When annotations are available for validation, we can efficiently optimize the hyperparameter~$\lambda$ in an automated fashion. For a variety of applications, we identify the optimal regularization weight~$\optimal$ for sets of 50 registration pairs. First, we investigate how~$\optimal$ differs across subject subpopulations and anatomical regions: we train \methodname on a subset of our T1w training data, and optimize~$\lambda$ separately for sets of ABIDE, GSP, MCIC, and UK Biobank (UKB) subjects at test time. With this single \methodname model, we also identify separate values of~$\optimal$ for a range of neuroanatomical regions. Second, we train \methodname on a subset of the multi-contrast image pairs and determine~$\optimal$ separately for T1w-to-T2w, T2w-to-T2w, and multi-flip-angle (multi-FA) registration tasks. Last, we analyze the extent to which~$\optimal$ differs between cross-sectional and longitudinal registration: we train \methodname on a combination of within-subject and cross-subject pairs from OASIS-2 and separately optimize~$\lambda$ for test pairs within and across subjects.

\begin{figure}[t]
  \centering
  \includegraphics[width=\textwidth]{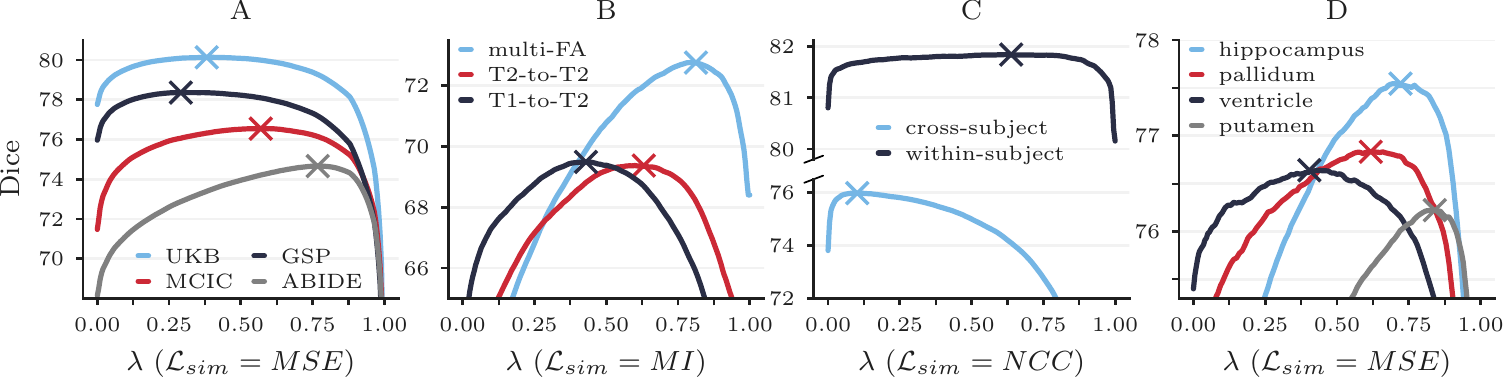}
  \caption{Registration accuracy across dataset subpopulations (A), image contrasts (B), tasks (C), and neuroanatomical regions (D). The cross markers indicate the optimal value~$\optimal$ as identified by automatic hyperparameter optimization.}
  \label{fig:downstream-optimization}
\end{figure}

\subpara{Results}
Figure~\ref{fig:downstream-optimization} shows that~$\optimal$ varies substantially across subpopulations, image contrasts, tasks, and anatomical regions. Importantly, in some cases using the~$\optimal$ computed for one subset of data on another results in considerably reduced accuracy. For example, using~$\optimal$ determined for GSP on ABIDE data decreases the maximum attainable Dice score by~$1.86 \pm 2.87$. We hypothesize that the observed variability in optimal hyperparameter values is caused by differences in image quality and anatomy between the datasets. Similarly, using the multi-FA~$\optimal$ for T1w-to-T2w registration and the within-subject~$\optimal$ for cross-subject registration causes the respective maximum Dice scores to drop by~$3.16 \pm 2.14$ and~$1.73 \pm 1.20$. Lastly, Figure~\ref{fig:downstream-optimization}D illustrates that the optimal~$\lambda$ value varies broadly across anatomical regions, suggesting that it is desirable to choose regularization weights depending on the downstream task and focus of a given study. In our experiments, automatic hyperparameter optimization takes just~$12.3 \pm 1.8$ seconds on average per test pair and requires 10 GB of memory, while interactive tuning requires only 2 GB. We emphasize that these metrics are influenced substantially by the size of the image data being registered.

\subsection{Experiment 4: Hypernetwork Size}

We also measure the importance of hypernetwork capacity for accurate representation of individually trained baseline models. 

\subpara{Setup}
We train separate \methodname models for three hypernetwork sizes: small (with 16, 16, 16, and 16 units per layer), medium (with 32, 32, 64, 64, and 64 units per layer), and large (with 32, 64, 64, 128, and 128 units per layer). We carry out these and all subsequent experiments using MSE for~$\Loss_{sim}$ and evaluate model accuracy against baselines results for 50 image pairs.

\subpara{Results}
Figure~\ref{fig:network-analysis}A shows that the capability of \methodname to match baseline registration accuracy increases with hypernetwork size. The large hypernetwork is appropriate for learning the effect of the regularization weight~$\lambda$ in 3D registration. Although the large hypernetwork contains approximately $7.6$ times more trainable weights than the small network, we find no substantial difference ($<0.4\%$) in total training or inference time across hypernetwork sizes, likely because the significant bottleneck is caused by convolutional operations.

\subsection{Experiment 5: Hyperparameter Sampling}

This experiment evaluates how different hyperparameter sampling methods affect \methodname accuracy. In previous tests, we observe that sampling regularization weights~$\lambda$ from a uniform distribution during \methodname training results in registration accuracy comparable to baseline models across most of the hyperparameter range, especially near~$\optimal$, but less comparable estimations very close to the boundaries~$\lambda\in\{0,1\}$, corresponding to similarity-only or regularization-only loss functions.

\subpara{Setup}
To investigate whether these boundaries can also be captured by \methodname, we over-sample the end-point values~$\{0,1\}$ of the hyperparameter~$\lambda$ at a fixed rate~$r$. We train and evaluate three separate models for different values of~$r$ (0.0, 0.2, and 0.8) and compare the final accuracy against baselines, to asses the influence of this rate on registration accuracy.

\subpara{Results}
\methodname models trained for large values of~$r$ closely match the expected registration accuracy at end-point values of~$\lambda$ but sacrifice registration accuracy across all values of~$\lambda$ (Figure~\ref{fig:network-analysis}B). For example, when training \methodname with~$r = 0.0$ (no over-sampling), the mean deviation from the baseline Dice is $0.08 \pm 0.26$ at~$\optimal$, compared to $2.96 \pm 1.57$ at~$\lambda \in \{0,1\}$. However, with~$r = 0.8$, the mean deviation from baseline Dice is~$0.87 \pm 0.40$ at~$\optimal$ and~$0.49 \pm 0.51$ at~$\lambda \in \{0,1\}$. We emphasize that over-sampling is only necessary to estimate appropriate representations at the extreme hyperparameter boundaries. As similarity-only or regularization-only loss functions are not desirable for the majority of applications, uniform sampling will suffice in most cases. Throughout all experiments presented in this study, we choose an intermediate value of~$r = 0.2$, which facilitates the most consistent matching of baseline registration accuracy for all values of~$\lambda$.

\subsection{Experiment 6: Alternative Models}
\label{sec:altexperiment}

While hypernetworks facilitate learning the effect of hyperparameters on registration networks, we also investigate the alternative \methodname strategy of adding an input to the standard registration network.

\subpara{Setup}
We train the \methodpre, \methodpost, and \methodfull architectures defined in Section~\ref{sec:altmethod} and compare the resulting registration accuracy and computational cost with the individual baseline models.

\begin{figure}[t]
  \centering
  \includegraphics[width=\textwidth]{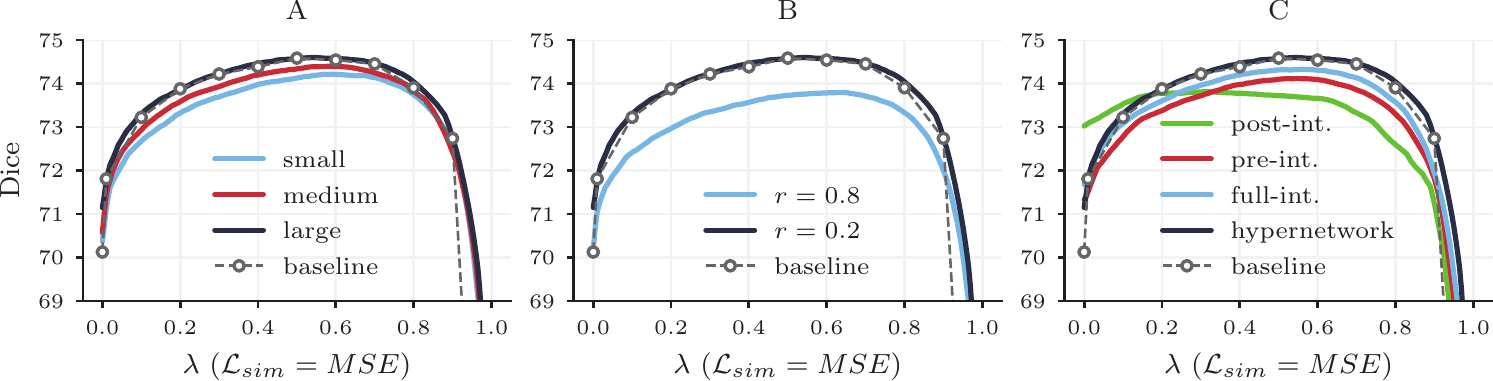}
  \caption{Analysis showing registration accuracy in terms of Dice overlap for \methodname models trained with different hypernetwork sizes (A), end-point sampling rates~$r$ (B), and \methodname strategies (C).}
  \label{fig:network-analysis}
\end{figure}

\subpara{Results}
None of the three alternative networks yield the accuracy achieved by baseline and hypernetwork-based \methodname models~(Figure~\ref{fig:network-analysis}C). While the \methodpre and \methodfull networks broadly encapsulates the effect of~$\lambda$, identifying the baseline~$\optimal$ with an average error of~$0.07 \pm 0.08$ and~$0.05 \pm 0.07$, respectively, they deviate from peak baseline accuracy by~$0.48 \pm 0.25$ and~$0.23 \pm 0.46$. The \methodpost network struggles to learn the accurate effect of~$\lambda$, deviating from the baseline~$\optimal$ by $0.19 \pm 0.12$ and peak accuracy by~$0.71 \pm 0.61$. The total train time for the \methodfull model is~$1.1\times$ longer than that of the hypernetwork-based \methodname, while the \methodpre and \methodpost models require~$1.8\times$ more time, likely due to the added convolutional operations in the network.


\section{Discussion and Conclusion}

The accuracy of learning-based deformable registration algorithms largely hinges on the choice of adequate hyperparameter values, which might differ substantially across registration targets, data types, model architectures, and loss implementations. Consequently, accurate and high-resolution hyperparameter search is an essential component of registration model development.

In this work, we present \methodname, a learning strategy for registration that eliminates the need to repeatedly train the same model with different hyperparameter values to evaluate their effect on performance. \methodname employs a hypernetwork that takes the desired hyperparameter values as input and predicts the corresponding parameters, or weights, of a registration network. We show that training a \textit{single} \methodname model is sufficient to capture the behavior of a range of baseline models individually optimized for different hyperparameter values. This enables \textit{precise} hyperparameter optimization at test-time, because the optimal value may be located between the limited number of discrete grid points evaluated by traditional approaches.

We explore two alternatives for choosing optimum regularization weights: one interactive, based jointly on image matching and visual smoothness, and one automatic, based on registration accuracy. The automatic method optimizes Dice overlap, which in itself does not take field regularity into account. We ensure that this parameterization yields regular deformations by analyzing voxel-wise Jacobian determinants. However, we emphasize that HyperMorph enables efficient optimization of hyperparameter values at test-time using any desired metric of choice.

\subpara{Function vs.\ Input Space} We explore two paradigms for learning the effect of registration hyperparameters on the deformation field: a hypernetwork-based function that returns an appropriate registration function given a hyperparameter value or a modification of the registration function to accept an addition hyperparameter value as input (\methodpre, \methodpost, or \methodfull). In the analysis, the latter approach under-performs in registration quality, and thus, modelling the effect of hyperparameters in this manner presents a more challenging optimization problem. We hypothesize that this effect could be due to the fact that the convolutional filters are fixed once training is complete, requiring them to perform a substantially more difficult task than simple registration. In contrast, the hypernetwork approach enables the convolutional filters to flexibly adapt to specific hyperparameter values, suggesting a more powerful mechanism. Further analysis of these effects is an interesting future direction but is beyond the scope of this work.

We emphasize that a hypernetwork is not the only effective mechanism for learning the effects of hyperparameters on registration networks, and we investigate this group of alternative architectures in an attempt to gain and provide insight across approaches. For example, parallel, independent work~\citep{mok2021} explores conditional registration networks. These learn regularization effects by leveraging instance normalization and employing an MLP to scale and shift hidden features as a function of the regularization weight~$\lambda$. This strategy is similar to the \methodfull implementation, suggesting another promising alternative strategy. The approach is also similar to hypernetwork-based \methodname since it employs an MLP to learn the hyperparameter effect, but it differs in how this MLP is coupled with the registration network. It is likely that with sufficient architectural optimization, both hypernetworks and specifically-designed conditional CNNs are powerful solutions for a variety of hyperparameter learning tasks.

\subpara{Computational efficiency} By exploiting the similarity of networks across a range of hyperparameter values, a single \methodname model uses weight-sharing to efficiently learn to estimate optimal deformation fields for arbitrary image pairs and \textit{any} hyperparameter value from a continuous interval. This enables fast, automated tuning of hyperparameters at test time and represents a substantial advantage over traditional search techniques: to identify an optimal configuration, these techniques typically optimize a number of registration networks across a sparse, discrete grid of hyperparameter values, which requires dramatically more compute and human time than \methodname.

\subpara{Initialization robustness}
Experiment 2 demonstrates that \methodname is substantially more robust to network weight initialization than individually trained networks, exhibiting 43 to 61\% reduced variability over training runs, likely because the combined hypernetwork and registration-network stack can take advantage of weight-sharing across a landscape of hyperparameter values.
This result further underlines the computational efficiency provided by \methodname, since traditional tuning approaches often resort to training models multiple times at each grid point to remove potential bias due to initialization variability.

\subpara{Test-time adaptation}
Existing registration models are often trained using a single hyperparameter value optimized globally for a set of validation data. However, the frequently overlooked reality is that hyperparameter optima can differ substantially across individual image pairs and applications, whereas most, if not all, registration-based analysis pipelines assume the existence of a single optimal hyperparameter value~\citep{patenaude2011bayesian,wang2012multi,fischl2012}. For example, a pair of images with very different anatomies would benefit from weak regularization, permitting warps of high non-linearity. This implies that learning-based methods capable of adapting hyperparameters on the fly are essential. We demonstrate that a single \methodname model enables rapid discovery of optimal hyperparameter values for different dataset subpopulations, image contrasts, registration tasks, and even individual anatomical regions, facilitating the future development of models that learn to estimate ideal hyperparameter values for individual registration pairs.

\subpara{Further work}
\methodname can be used with hyperparameters beyond those evaluated in this work. For example, it could be applied to the number of bins in the MI metric, the choice of form of the regularization term~$\lambda_{reg}$, the hyperparameters used in the regularization term(s), the level of dropout, or even architectural hyperparameters, similarly to the SMASH method~\citep{brock2017}. However, the effects of certain hyperparameters, especially those related to model architecture, might be substantially more difficult for a hypernetwork to learn. Additionally, for \methodname to learn the effects of some optimization-specific parameters, like learning rate and batch size, it would likely require substantial modifications.

\begin{wrapfigure}{R}{0.5\textwidth}
  \centering
  \includegraphics[width=0.4\textwidth]{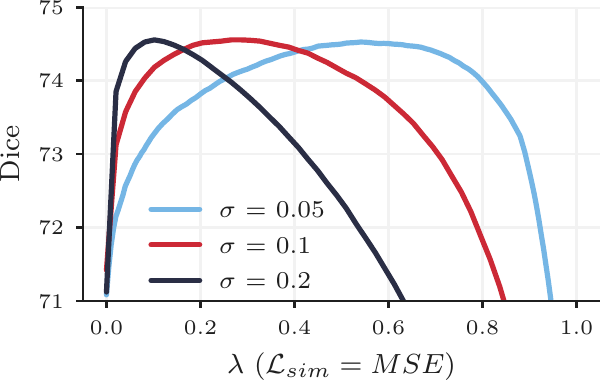}
  \caption{Registration accuracy (Dice) achieved by \methodname models trained for different values of estimated image noise~$\sigma^{-2}$.}
  \label{fig:sigma-analysis}
\end{wrapfigure}

The identification of~$\optimal$ for different brain regions promotes a potential future direction of estimating a spatially-varying field of regularization hyperparameters for simultaneously optimal registration of all anatomical structures.
Additionally, while we evaluate \methodname for one and two hyperparameters, we expect this strategy to readily adapt to more hyperparameters and are eager to explore hypernetworks in this context, in which grid search is impractical. We are also interested in investigating how the benefits of implicit weight-sharing in hypernetworks might differ across categories of loss hyperparameters.

We also plan to expand this work by exploring more complex distributions of~$p(\Lambda)$ and how they affect hyperparameter search. For example, in registration formulations where the image similarity term is re-weighted by estimated image noise~$\sigma^{-2}$, the range of the hyperparameter space that should be searched can vary substantially. With a suboptimal choice of~$\sigma$, a grid search is often even more challenging, as the range of hyperparameter values that perform well can be very narrow. In a preliminary experiment, we found that \methodname performed well for a variety of noise estimates~$\sigma$, even with a uniform distribution~$p(\lambda) = \mathcal{U}(0, 1)$ used throughout our experiments (Figure~\ref{fig:sigma-analysis}). However, the result also simultaneously highlights the more dramatic Dice score sensitivity to hyperparameter choice for some~$\sigma$ values, suggesting that non-uniform distributions might lead to even better \methodname performance.

\subpara{Conclusion}
We believe \methodname has the potential to drastically alleviate the burden of retraining networks with different hyperparameter values, thereby enabling efficient development of finely optimized models for image registration. While the training strategy described in this paper is well-suited for tuning a visually-driven workflow like image registration, the technique can be used to improve other applications within and beyond the domain of medical imaging analysis.

\acks{Support for this research was provided in part by the BRAIN Initiative Cell Census Network (U01 MH117023), the National Institute for Biomedical Imaging and Bioengineering (P41 EB015896, 1R01 EB023281, R01 EB006758, R21 EB018907, R01 EB019956, P41 EB030006), the National Institute on Aging (1R56 AG064027, 1R01 AG064027, 5R01 AG008122, R01 AG016495, 1R01 AG070988), the National Institute of Mental Health (R01 MH123195, R01 MH121885, 1RF1 MH123195), the National Institute for Neurological Disorders and Stroke (R01 NS0525851, R21 NS072652, R01 NS070963, R01 NS083534, 5U01 NS086625, 5U24 NS10059103, R01 NS105820), the NIH Blueprint for Neuroscience Research (5U01 MH093765), the multi-institutional Human Connectome Project, the National Institute of Child Health and Human Development (K99 HD101553), and the Wistron Corporation. This research was made possible through resources provided by Shared Instrumentation Grants 1S10 RR023401, 1S10 RR019307, and 1S10 RR023043.}

\ethics{The work follows the highest ethical standards in conducting research and writing the manuscript. All models were trained using publicly available data, with the exception of the FSM data. We believe that \methodname can substantially improve the use of image registration in medical image analysis, and we use the UK Biobank data in accordance with this interest in public health.}

\coi{Bruce Fischl has a financial interest in CorticoMetrics, and his interests are reviewed and managed by Massachusetts General Hospital and Mass General Brigham in accordance with their conflict of interest policies.}

\bibliography{references}

\newpage

\renewcommand{\thefigure}{S\arabic{figure}}
\setcounter{figure}{0}

\begin{figure}[t]
  \centering
  \includegraphics[width=\textwidth]{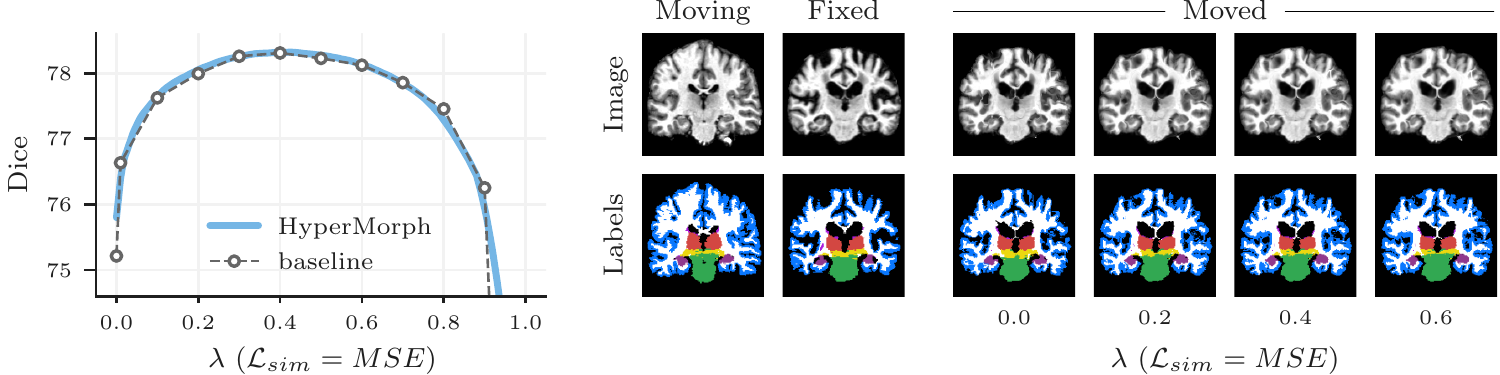}
  \caption{Left: mean Dice scores achieved by a single HyperMorph model and baselines evaluated on the held-out, manually-labeled Buckner40 dataset. Right: image and label-based qualitative changes in HyperMorph alignment across different regularization weights for a given subject pair.}
  \label{fig:held-out}
\end{figure}

\begin{figure}[t]
  \centering
  \includegraphics[width=\textwidth]{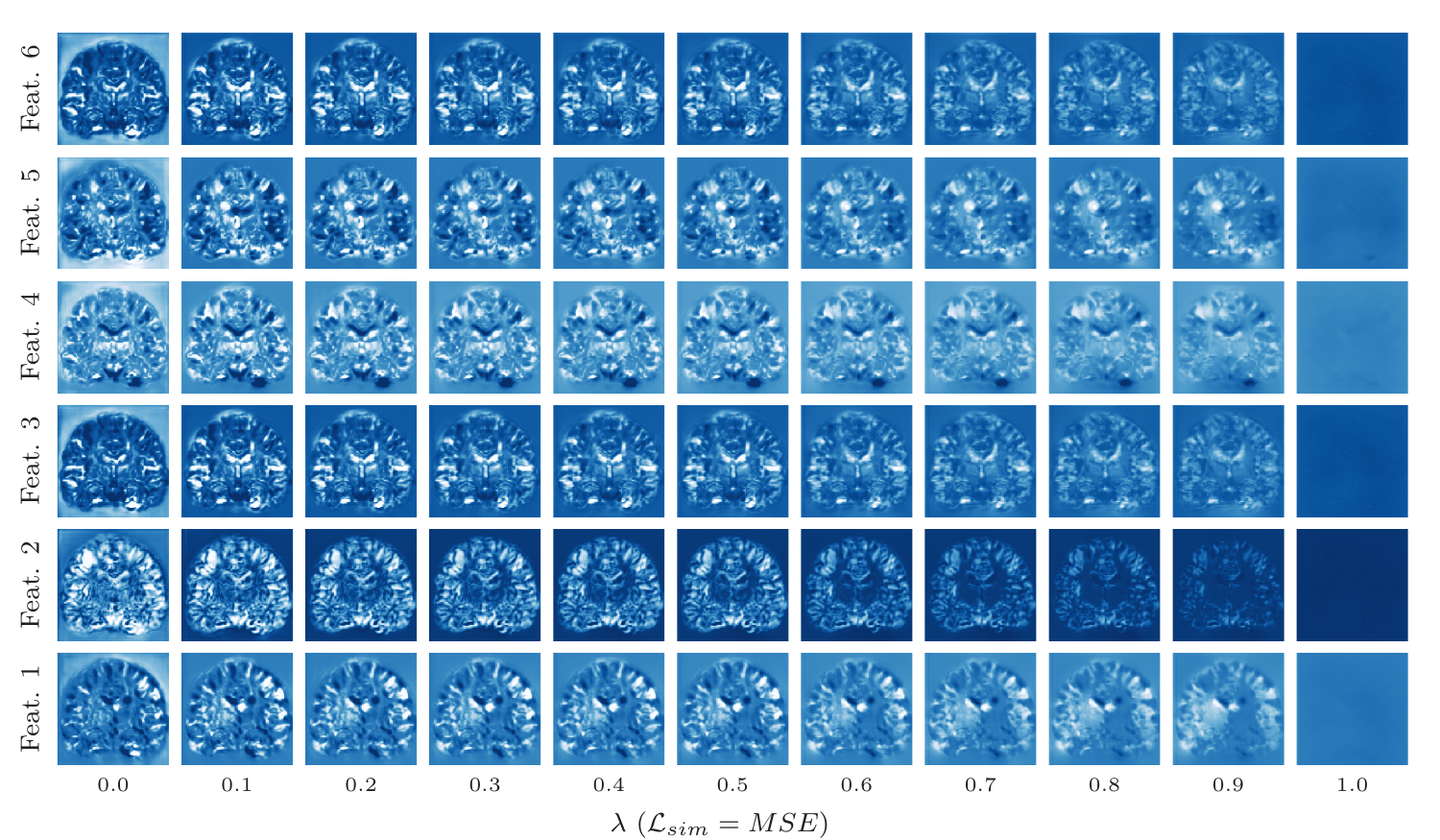}
  \caption{Changes in feature activations of the final HyperMorph U-Net layer across different values for~$\lambda$.}
  \label{fig:supplementary-features}
\end{figure}

\end{document}